\documentclass[lettersize,journal]{IEEEtran}
\usepackage{amsmath,amsfonts}
\usepackage{algorithmic}
\usepackage{algorithm}
\usepackage{amsthm}
\usepackage{amssymb}
\usepackage{array}
\usepackage[caption=false,font=normalsize,labelfont=sf,textfont=sf]{subfig}
\usepackage{textcomp}
\usepackage{stfloats}
\usepackage{bm}
\usepackage{url}
\usepackage{verbatim}
\usepackage{graphicx}
\usepackage{cite}
\usepackage{booktabs}
\usepackage{multirow}
\usepackage{doi}
\begin{document}

\title{Extremely low-bitrate Image Compression Semantically Disentangled by LMMs from a Human Perception Perspective}

\author{Juan Song, Lijie Yang, Mingtao Feng$^\ddagger$ 

\thanks{Juan Song, Lijie Yang, Mingtao Feng are with Xidian University, Xi’an 710071, China (Email: songjuan@mail.xidian.edu.cn; 
 23031212033@stu.xidian.edu.cn; 
mintfeng@hnu.edu.cn; 
)} }

\markboth{IEEE Transactions on Circuits and Systems for Video Technology}%
{Shell \MakeLowercase{\textit{et al.}}: A Sample Article Using IEEEtran.cls for IEEE Journals}


\maketitle

\begin{abstract}
It remains a significant challenge to compress images at extremely low bitrate while achieving both semantic consistency and high perceptual quality.
Inspired by human progressive perception mechanism, we propose a Semantically Disentangled Image Compression framework (SEDIC) in this paper. Initially, an extremely compressed reference image is obtained through a learned image encoder. Then we leverage LMMs to extract essential semantic components, including overall descriptions, object detailed description, and semantic segmentation masks. We propose a training-free Object Restoration model with Attention Guidance (ORAG) built on pre-trained ControlNet to restore object details conditioned by object-level text descriptions and semantic masks. 
Based on the proposed ORAG, we design a multistage semantic image decoder to progressively restore the details object by object, starting from the extremely compressed reference image, ultimately generating high-quality and high-fidelity reconstructions.
Experimental results demonstrate that SEDIC significantly outperforms state-of-the-art approaches, achieving superior perceptual quality and semantic consistency at extremely low-bitrates ($\le$ 0.05 bpp).
\end{abstract}


\section{Introduction}
With the rapid proliferation of visual data, the demand for extremely low-bitrate image compression has become increasingly critical. By compressing images to as little as one-thousandth of their original size, such techniques effectively reduce storage and transmission costs, especially in bandwidth-constrained scenarios. However, reconstructing high-fidelity images from these highly compressed representations remains a significant challenge due to severe information loss. Consequently, designing advanced compression strategies that balance fidelity and perceptual quality at extremely low-bitrates has emerged as a key research focus.

Traditional compression codecs, e.g., JPEG \cite{wallace1992jpeg} and VVC \cite{vvc}, are constrained to use large quantization steps in such scenarios, inevitably leading to severe blurring and blocking artifacts. Despite the superior rate-distortion (R-D) performance of learning-based compression techniques \cite{2016endtoend,2018-context_hyperprior,2018VAE,2020CC_LRP,2023-TCSVT-GAN-loss} that follow the Variational Autoencoders (VAEs), these methods produce blurry images at extremely low-bitrates, due to the reliance on optimization of pixel-oriented distortion metrics measured by the Mean Square Error (MSE) and Structural Similarity Index Measure (MS-SSIM), which are not fully consistent with humans’ perceptual quality.

\begin{figure}[t!]
    \centering
    \includegraphics[width=1.0\linewidth]{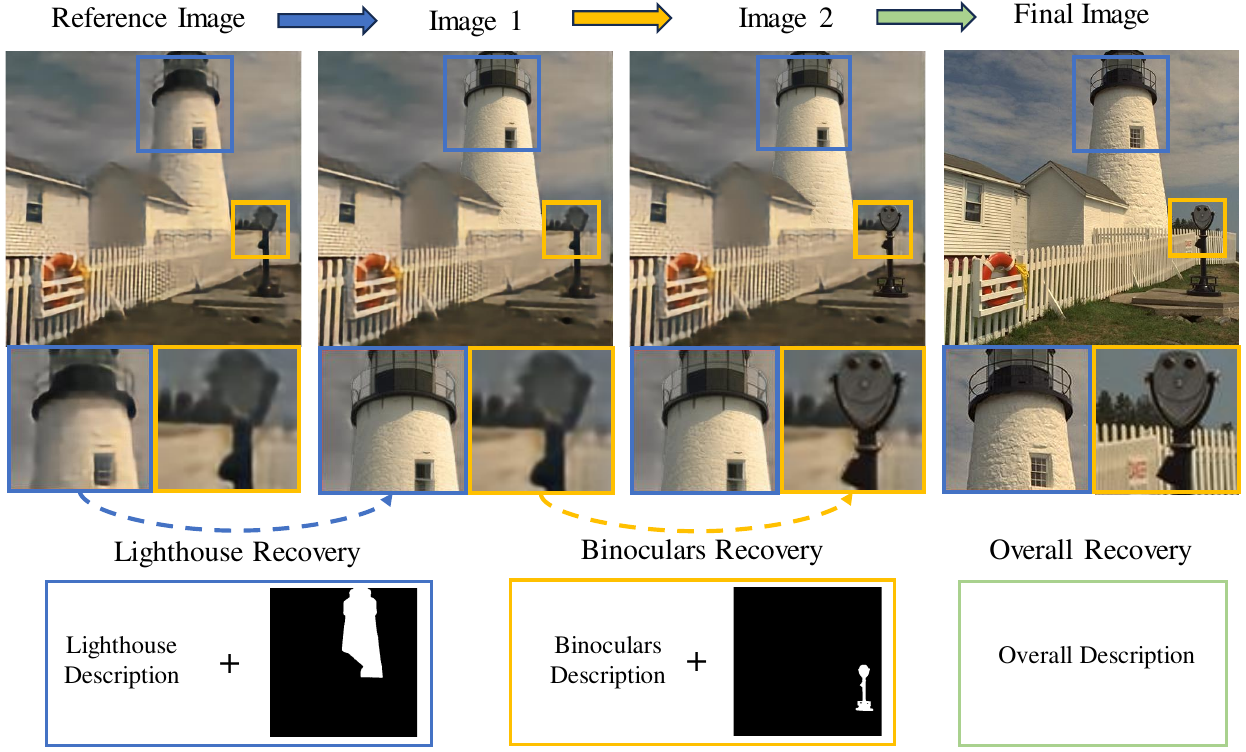}
  
   \caption{Starting from the extremely compressed reference image, our proposed ORAG firstly progressively restores details object by object conditioned by object descriptions and semantic masks. Finally, the overall description is used to enhance the overall perceptual quality. 
   }
   \vspace{-3mm}
    \label{fig:object_restoration}
   
\end{figure}

To address this issue, Generative Image Compression (GIC)\cite{2023-TCSVT-ULCompress,GLC-image} begins to prioritize semantic consistency with the reference image over preserving pixel-level fidelity. Generative adversarial networks (GANs) are used as decoders, generating impressive results in terms of perception quality. Diffusion models further advance GIC by reconstructing images with richer visual details, albeit at the cost of some fidelity to the original image.

The emergence of large multimodal (LMM) models, e.g., GPT-4 Vision \cite{gpt4-vision} has introduced new paradigms for extremely low-bitrate image compression, which encode the images into compact semantic representations such as text, sketch map \cite{lei2023text+sketch}.
Pre-trained text-to-image Stable Diffusion models \cite{stablediffusion} are employed in the decoder constrained by transmitted semantic representations to produce reconstructions with high perceptual quality. However, current LMM models still struggle to generate complicated prompts involving adequate details in images, resulting in semantic detail inconsistency with the original image.
That motivates us to think about the questions: \textit{How to disentangle the image into compact semantic representations leveraging the capacity of LMMs?  How can we maintain the trade-off between perception and semantic consistency under extremely low-bitrate constraints?}

Research \cite{2005-coarse,2008-timecourse} in visual cognition and neuroscience suggest that  human perception is usually progressive. 
Our eyes tend to firstly capture an overview of the image at a glance, which tends to be unfocused and blurred with low quality. Subsequently, by directly focusing on the objects of interest, our eyes acquire detailed and high-resolution information regarding the objects. Inspired by this biological phenomenon, we design a novel SEmantically Disentangled Image Compression (SEDIC) framework to imitate this progressive perception. Initially, an extremely compressed reference image is obtained through a learned image encoder. Then, we leverage LMMs to extract essential semantic information regarding objects of interest, including overall description, object-detailed description, and semantic segmentation masks. We propose an training-free Object Restoration model with Attention Guidance (ORAG) built on pre-trained ControlNet\cite{2023-controlnet} to restore object details conditioned by object-level text descriptions and semantic masks. Based on ORAG, we design a multistage semantic image decoder. Starting from the extremely compressed reference image, as illustrated in Figure \ref{fig:object_restoration}, the image decoder progressively restores the details object by object, ultimately generating high-quality and high-fidelity reconstructions.  
The contributions are summarized as follows.

\begin{itemize}
\item We propose a semantically disentangled image compression framework by leveraging the great capacity of LMMs to disentangle the image into compact semantic representations, including an extremely compressed reference image, semantic masks, overall and object-level text descriptions. In particular, semantic masks can provide semantic alignment with the object description in the reference image to facilitate subsequent object restoration.

\item We propose an Object Restoration model with Attention Guidance (ORAG) to restore object details conditioned by object detailed descriptions and segmentation masks. Based on ORAG, we design a multi-stage semantic decoder that performs restoration object-by-object progressively, starting from the extremely compressed reference image, ultimately generating high-quality and high-fidelity reconstructions.

\item Both qualitative and quantitative results demonstrate that proposed SEDIC achieves significant improvements compared to SOTA codecs in terms of perceptual quality 
 metrics at extremely low-bitrates ($\le$ 0.05bpp).
\end{itemize}

\section{Related Works}

\subsection{Extremely-low Bitrate Image Compression.}
The majority of extremely low bitrate image compression approaches fall into the fields of generative image compression, which leverage GAN or Diffusion models to achieve perceptually good reconstructions. HiFiC \cite{2020-Hific} and Muckley et al. \cite{muckley2023improving} demonstrated the effectiveness of the GAN-based decoder for human perception by introducing a divergence term typically in the form of an adversarial discriminator. Yang et al. \cite{2024CDC} replaced the decoder network with a diffusion model which is conditioned by the transmitted latent variables. Diffusion models have also empowered the breakthrough in text-to-image generation models, enabling to create realistic images given text descriptions. Recent works explore compression of images into extremely compressed semantic information, such as text \cite{pan2022extreme}, sketch map \cite{lei2023text+sketch}, or vector-quantized image representations \cite{2024-ICLR-PerCo}. which are decoded and used as the conditional input for image generation.
Despite these advantages, they still struggle to achieve a satisfactory trade-off between the consistency and perceptual quality at such low bitrates.

\subsection{Large Multimodal Models.}
Large Multimodal Models (LMMs) have demonstrated remarkable reasoning and understanding capabilities in vision-language tasks, including visual question answering \cite{gpt4-vision,2023ferret-QA,2024visual-QA} and document reasoning \cite{2024cogagent-document-reasoning,2024textmonkey-document-reasoning}. In particular, Multimodal Large Language Models (MLLMs) like GPT-4 Vision \cite{gpt4-vision} enable rich visual-textual interaction by generating detailed image descriptions and supporting joint image-text inputs. Complementing these, vision-centric models such as Grounding DINO \cite{2023-groundingDino} and Segment Anything Model (SAM) \cite{2023-SAM} provide open-vocabulary object detection and high-quality mask generation, further enhancing semantic understanding. Motivated by their great comprehensive capabilities, recent work has explored LMM to compress images into semantic representations. SDComp \cite{liu2024-SDComp} leveraged LMMs to perform importance ranking and semantic coding for downstream machine vision tasks; Murai et al. \cite{murai-2024-lmm} generate image captions and compress them within a single LMM model. Our work is most related to MISC \cite{li2024-misc} which encodes images into text, spatial maps, and an extremely compressed image. 
However, spatial maps cannot provide precise spatial positions to semantically align text descriptions with objects in the reference image.
In addition, MISC restored each object conditioned on previously restored objects in the pixel domain, which may introduce noticeable boundaries between spatial maps.
The above drawbacks lead to the fact that the object information guides the diffusion model in a less significant way.

\subsection{Controllable Image Generation.}
Diffusion models have garnered significant attention due to their powerful generative capability. Text-to-image generation \cite{2023-Rich-text-to-image} is one of the most popular applications, which aims to generate high-quality images aligned with given text prompts. Additionally, several studies \cite{ACMMM-magic,2023-controllableface,2024-loosecontrol,2023-controlnet}, e.g. ControlNet, further augmented controllability by adding spatially localized input conditions, e.g., edges, depth, segmentation and human pose, to a pre-trained text-to-image diffusion model. Based on ControlNet\cite{2023-controlnet}, Lin et al.\cite{2023Diffbir} proposed IRControlNet that leverages text-to-image diffusion prior for realistic image restoration. Li et al. \cite{2023-MuLan} proposed a multimodal LLM agent (MuLan) that utilized a training-free multimodal-LLM agent to progressively generate objects with feedback control. 
 We aim to exploit controllable image generation techniques for object-level semantic decoding, thereby maintaining high visual fidelity and perception quality.

\begin{figure*}[t!]
    \centering
    \includegraphics[width=1.0\linewidth]{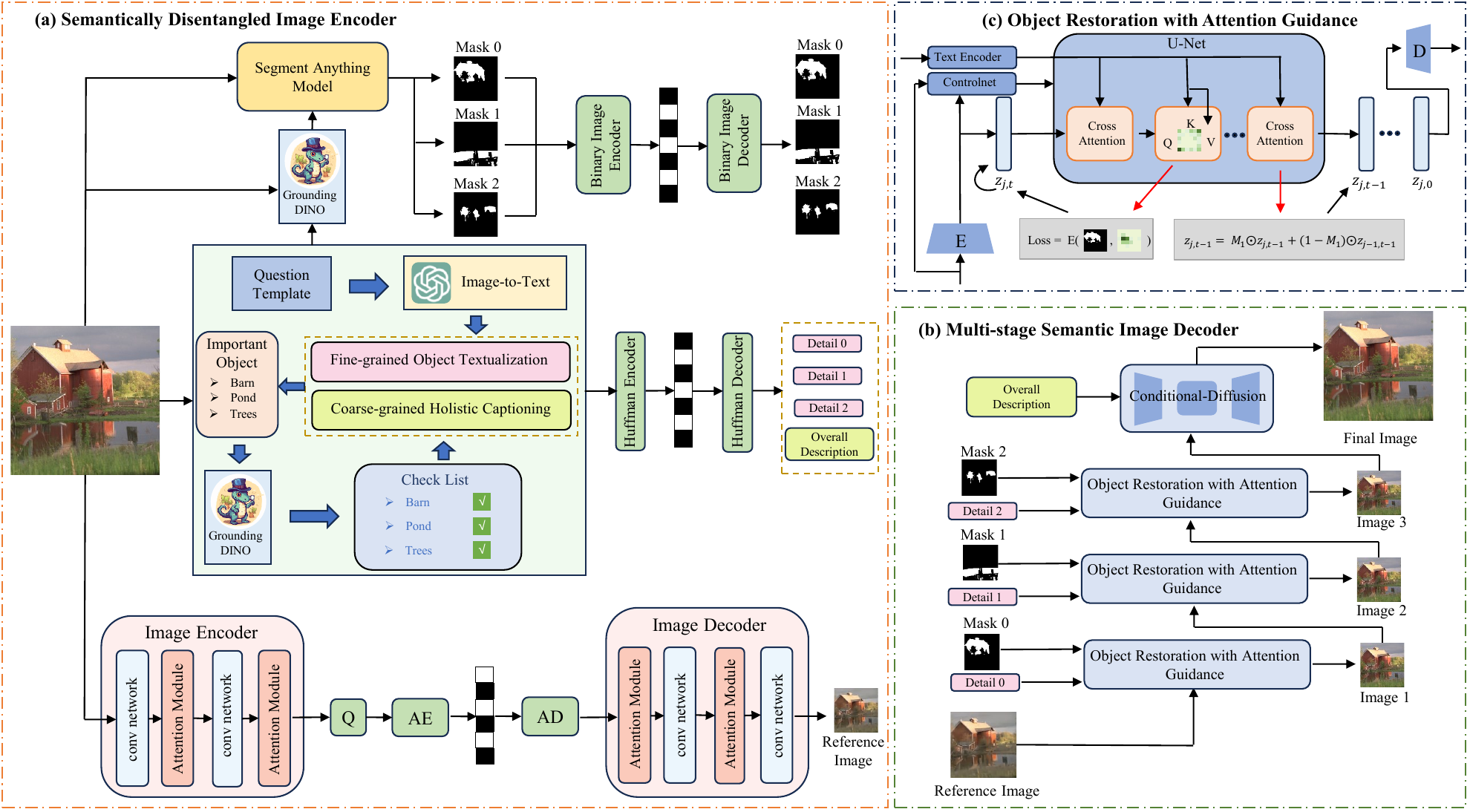}
    \caption{Overall framework of SEDIC. (a) Semantically Disentangled image encoder consists of an image textualization encoder to extract overall and object-level detailed descriptions, a semantic mask encoder, and an image encoder to obtain an extremely compressed reference image. (b) Multi-stage Semantic Image Decoder consists of several Object Restoration models with Attention Guidance (ORAG) to restore object details and a conditional text-to-image diffusion model to restore the entire image. (c) The ORAG model restores the object details given object text descriptions and  semantic masks. 
    }
    \label{fig:SIC}
   
\end{figure*}

\begin{figure}[t!]
    \centering
    
    \includegraphics[width=1\linewidth]{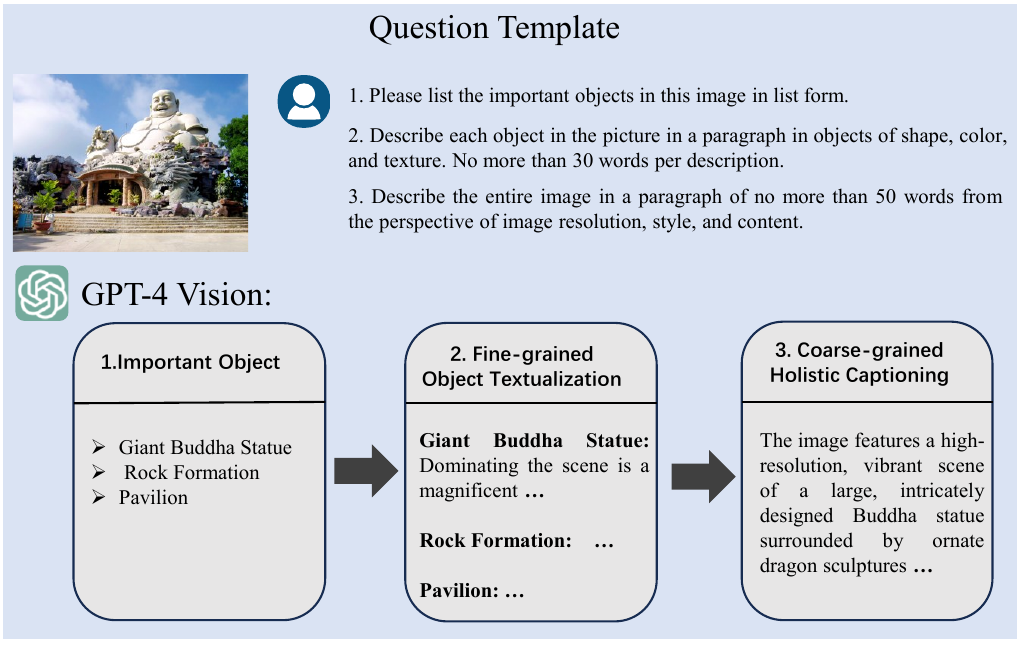}
    \caption{Question template designed to guide GPT-4 Vision in image-to-text encoding.The template comprises three stages: (1) object listing, (2) fine-grained object-level textualization, and (3) holistic image-level captioning.}
    
    \label{fig:Question_Template}
\end{figure}

\section{Methodology}
In this section, we propose a semantically disentangled image compression framework, as illustrated in Figure \ref{fig:SIC}. Semantically Disentangled Image Encoder, consisting of LMM models, disentangles images into holistic and object-grained text descriptions, semantic masks and an extremely low-bitrate compressed image. 
Multi-stage Semantic Image Decoder, composed of ORAG models and a conditional diffusion model, progressively restores the image from object-level to global structure conditioned by semantic components.

\subsection{Semantically Disentangled Image Encoder}

\noindent \textbf{Image Textualization Encoder.}
Text description is the compact semantic representation of the image. Existing image-to-text based coders only used a brief and holistic text description lacking
details to guide generative decoders. That results in low fidelity with ground truth, although satisfactory perception quality is achieved \cite{lei2023text+sketch,pan2022extreme}. Inspired by recent advancements in image captioning\cite{pi2024image}, we design an Image Textualization Encoder that generates detailed descriptions of significant objects along with holistic descriptions of the entire image. This process operates in two stages: fine-grained object textualization and coarse-grained holistic captioning. 
\textbf{Fine-grained Object Textualization}. We utilize the powerful visual understanding capabilities of the most advanced GPT-4 Vision \cite{gpt4-vision} model to generate fine-grained object-level descriptions focusing on object attributes such as shape, color, texture. The image is encoded into Object Name $Textn_j\left( \leqslant l_n\ words \right) $ and Object Details $Textd_j$:  $\left( \leqslant l_d\ words \right)$ ($j=0,1,2...,J$),where J denotes the number of significant objects.
According to visual memory research\cite{alvarez2004capacity}, the capacity of visual memory depends on the number of objects and the visual information load, with an upper limit of 4 or 5 objects. Considering the visual memory capacity and extremely-low bitrate requirement(more objects higher bitrate), we set the upper limit of $J$ to 3.This setting ensures that essential objects are restored, balancing image compression efficiency and computational complexity.
\textbf{Coarse-grained Holistic Captioning}. 
Besides object-level descriptions, we also employ GPT-4 Vision model to produce an overall description of the image $ Text_{all} \left( \leqslant l_{all}\ words \right) $, summarizing broader aspects such as resolution, content and style. Although lacking detailed visual information, overall descriptions include primary objects and contextual information essential to preserve global coherence during reconstruction. The combination of detailed object descriptions and holistic captions facilitates the restoration of texture details and overall perceptual quality. Finally, we employ Huffman coding to losslessly compress text information {$Textd$, $Text_{all}$} at the minimum bitrate cost and transmit them to the decoder. 

Figure \ref{fig:Question_Template} illustrates the question template used in the MLLM model, GPT-4 Vision, for image textualization encoding. Through the prompts within the template, we guide GPT-4 Vision to generate Important Object listings, object-level descriptions, and global image captions. The word length for object descriptions, denoted as $l_d$, is dynamically adjusted based on the compression level, with a maximum limit of 30 words. Similarly, the word length for the overall image caption, denoted as $l_{\text{all}}$, is adjusted in accordance with the compression level, with a maximum of 50 words. This dynamic adjustment mechanism of word lengths ensures the system's flexibility, enabling it to adapt to varying bitrate requirements while maintaining a balance between bitrate and reconstruction quality.

Even the most powerful MLLMs, such as GPT4-Vision, suffer from the hallucination issue. It may generate descriptions of objects that do not exist in the image. To address this issue, we utilize Grounding DINO \cite{2023-groundingDino}, an open-world object detector with robust zero-shot detection capabilities, to verify whether each object in the descriptions is detected in the image. Any hallucinated object phrases, which are not found in the image, are tagged as "Hallucination" and removed from the text descriptions. 

\noindent \textbf{Semantic Mask Encoder.}
Text descriptions lack the ability to convey the precise spatial relationships between objects needed in image reconstruction. We propose a Semantic Mask Encoder that generates precise semantic segmentation masks given the object name $Textn$, to provide precise spatial information and edge contours for each object. Compared to sketch maps\cite{lei2023text+sketch} or spatial maps\cite{li2024-misc}, semantic segmentation masks provide a more effective way by semantically aligning text descriptions with objects in the reference image. This alignment facilitates subsequent object restoration during the decoding process.

The SAM model \cite{2023-SAM} is an open-world segmentation model capable of isolating any object within an image given appropriate prompts, e.g., points, boxes. However, SAM cannot directly identify masked objects given text inputs. We combine SAM with Grounding DINO \cite{2023-groundingDino} to support text input about the object. First, we input the  Object Name $Textn$ into Grounding DINO to obtain the object’s bounding boxes, and then pass them to SAM to generate the semantic segmentation mask.
The semantic mask for each object, as a form of binary image, represents pixels in two distinct states—typically black and white. Some binary image compression methods, e.g. JBIG2 \cite{2000-jbig2}, runlength coding \cite{1966-runlength-encoding}, can be applied to further losslessly compress the semantic masks.

\noindent \textbf{Image Encoder.}
While text descriptions and segmentation masks offer semantic and spatial cues, they are insufficient to fully reconstruct image details such as structure and color nuance \cite{lei2023text+sketch}. To address this, we introduce an extremely compressed reference image that preserves coarse structure and color information, serving as the starting point for multi-stage semantic decoding.

To extremely compress a reference image at full resolution, we retrained the existing deep learning-based image compression methods, such as the cheng2020-attn model in the learned image compression library CompressAI \cite{2020compressai}. Given an input image $I$, a pair of latent $y = g_a(I)$ and hyper-latent $ z = h_a(y)$ is computed. The quantized hyper-latent $\hat{z} = Q (z)$ is modeled and entropy coded with a learned factorized prior. The latent $y$ is modeled with a factorized Gaussian distribution $p(y|\hat{z} )=\mathcal{N}(\mu ,diag(\sigma ))$ whose parameter is given by the hyper-decoder $(\mu ,\sigma )=h_{s}(\hat{z} )$. The quantized version of the latent $\hat{y}  = Q(y -\mu ) + \mu $ is then entropy coded and passed through decoder $g_s$ to derive reconstructed image $\tilde{I}_0   = g_s(\hat{y})$. The loss function $\mathcal{L}$  of end-to-end training is formulated as,

\begin{equation}
\mathcal{L} = R(\hat{y}) + R(\hat{z}) + \lambda \cdot D(I, \tilde{I}_0)
\end{equation}

where $\lambda$ balances bitrate and distortion. Adjusting $\lambda$ enables control over compression ratio. Our framework is compatible with any learned image compression method.

\begin{algorithm}[t]
    
    \fontsize{9pt}{10pt}\selectfont
    \renewcommand{\algorithmicrequire}{\textbf{Input:}}
    \renewcommand{\algorithmicensure}{\textbf{Output:}}
    \caption{Multi-stage Semantic Image Decoding}
    \label{alg1}
    \begin{algorithmic}[1]
        \REQUIRE{Reference image $\tilde{I}_0$ , text description $Text_{all}$, $Textd$, semantic mask $M$, diffusion steps $T$, attention guidance timestep threshold $T'$,number of objects $J$, the CLIP text encoder, the fixed VAE encoder $\varepsilon(\bullet )$, the fixed VAE decoder $\mathcal{D(\bullet )} $,the pretrained ControlNet.}
        \ENSURE{Final Reconstructed Image $\tilde{I}_F$.}
        
    \FOR{$j=0:J$ }
    \STATE {$ z_{j,T} \sim \mathcal{N}  (0,\mathrm{\mathbf{I} })$;} \STATE{
        $cf_j = \varepsilon (I_{j} )$;
        }
     \IF{ $j<{J} $ }
           \STATE{
          $ctd_j=CLIP(Textd_j)$;
           }
               \FOR{$t=T:0$ }
        \IF{ $t>{T}' $ }
           \STATE{
           $z_{j,t} = z_{j,t}- \eta \cdot \nabla_{z_{j,t}}E(A,M_j,k)$;
           }
        \ENDIF
        \STATE{
        $z_{j,t-1}=  ControlNet(z_{j,t},ctd_j,t,cf_j) $;
            }
         \STATE{
         $ z_{j, (t-1)} = M_j \odot z_{j, (t-1)} + (1 - M_j) \odot z_{(j-1), (t-1)}$;
         }
        \ENDFOR
    \ELSE
    {
     \STATE{$ctd_j = CLIP(Text_{all});$}
         \FOR{$t=T:0$ }
          
       \STATE{
        $z_{j,t-1}=  ControlNet(z_{j,t},ctd_j,t,cf_j) $;
            }
        \ENDFOR
    }
    \ENDIF

        \STATE{
        $\tilde{I}_{j+1}=\mathcal{D}(z_{j,0})$;
        }
     \ENDFOR
   \STATE{return  $\tilde{I}_F=\tilde{I}_{J+1}$ }           
    \end{algorithmic}

\end{algorithm}

\subsection{Multi-stage Semantic Image Decoder}

We develop a multi-stage semantic image decoder that is implemented progressively starting from fine-grained object-level restoration to holistic image restoration, ultimately generating high-quality reconstructions that are highly consistent with the original images. This decoder leverages the capability of controllable diffusion models to restore adequate details constrained by the extremely compressed reference image, text descriptions and semantic masks. 
Specifically, we design a training-free Object Restoration model with Attention Guidance (ORAG) built on pre-trained ControlNet \cite{2023-controlnet}, which restores one object per stage, conditioned by object descriptions and semantic masks. Inspired by work \cite{2023-MuLan,chen2024training}, we integrate backward attention guidance into ORAG to ensure that the generated object details given by object description $Textd$ are accurately positioned within the mask region $M$. The complete procedure is listed in Algorithm \ref{alg1} and described as follows. 

\textbf{Condition Encoding}. In each stage, we utilize the fixed VAE encoder $\varepsilon(\bullet )$ to encode the reconstructed reference image $I_j$ into the latent space: $cf_j = \varepsilon(I_j )$. In addition, CLIP text encoder, a pre-trained model that provides a shared text-image embedding space, is utilized to produce the textual representations and inject them into the cross-attention layers of the denoising U-Net. 

\begin{figure*}[t!]
    \centering
    \includegraphics[width=1.0\linewidth]{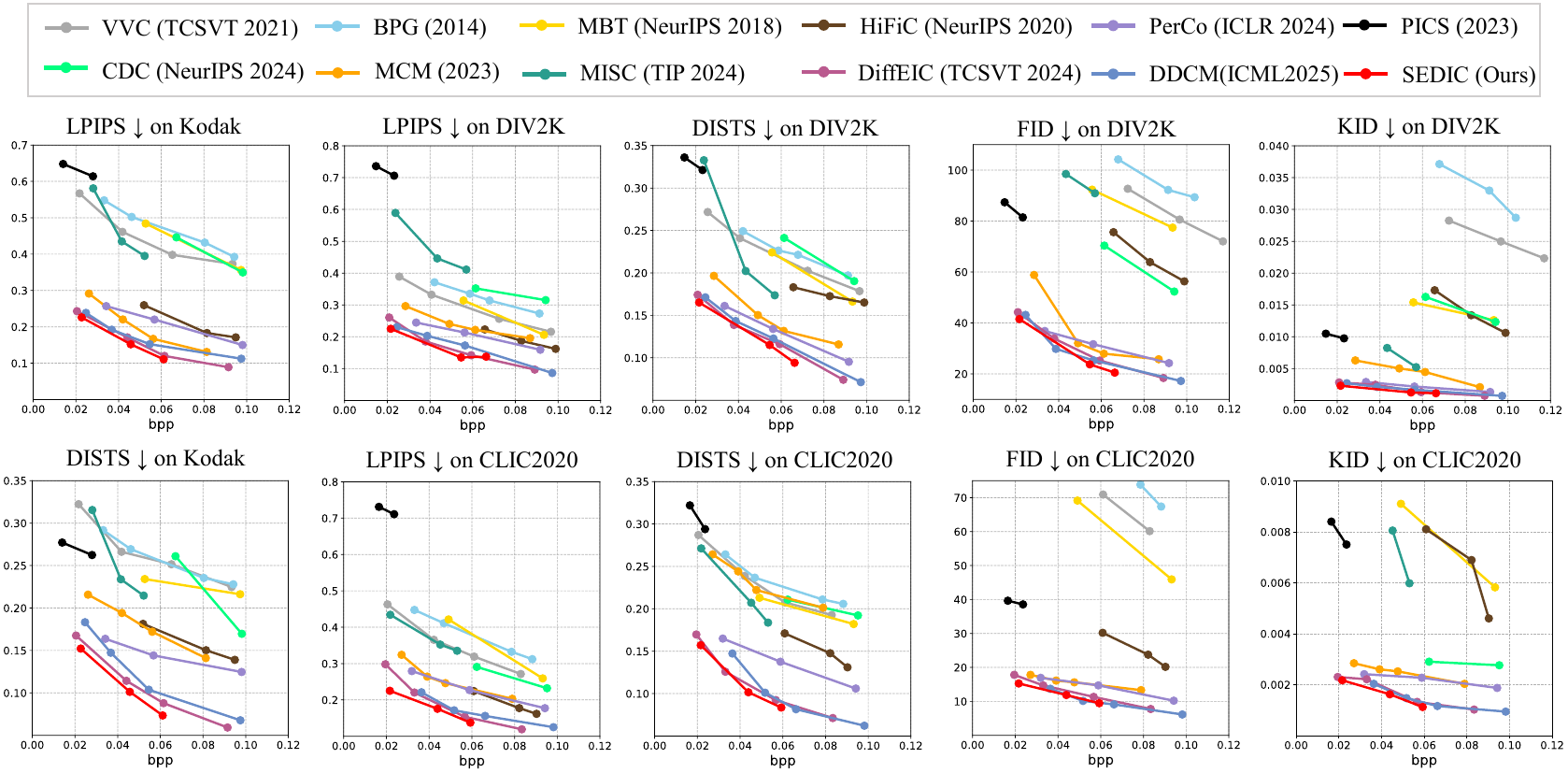}
    \vspace{-5mm}
    \caption{ Quantitative comparisons with SOTA methods in terms of perceptual quality (LPIPS$\downarrow$ / DISTS$\downarrow$ / FID$\downarrow$/ KID$\downarrow$) on Kodak \cite{kodak},
DIV2K validation \cite{DIV2K}, and CLIC2020 \cite{clic2020} datasets.
}
    \label{fig:result_image}
    \vspace{-3mm}
\end{figure*}

\textbf{Object Restoration with Attention Guidance.}
Given the object text description  $Textd_j$ and semantic mask $M_j$ of object $j$, our proposed training-free ORAG restores the object details in the reference image $I_j$ and ensures the restored object details will be correctly located within $M_j$. 
A natural and intuitive approach to achieve this in diffusion models is to guide the generation of the cross-attention map for objects, thereby establishing strong correlations between text descriptions and object semantic masks.
As illustrated in Figure \ref{fig:SIC}(c), our ORAG introduces backward guidance, which manipulates the cross-attention map under the guidance of the mask to maximize the relevance within the mask region. 
Specifically, let $A_{m,k}$ denote the cross-attention map which associates each spatial location $m$ of the immediate feature in the denoising network to token $k$ that describes object $j$ in the prompt $Textd_j$. Larger values in $A_{m,k}$ indicate a higher likelihood that the description is situated at that spatial location. 
The attention map is biased by introducing an energy function

\begin{equation}
    E\left(\boldsymbol{A}, \boldsymbol{M}_{j}, k \right)=\left(1-\frac{\sum_{m \in M_{j}} \boldsymbol{A}_{m, k}}{\sum_{m} \boldsymbol{A}_{m, k}}\right)^{2}
    \label{eq:ef}
\end{equation}

where $\sum_{m \in M_{j}}$ denotes the summation over the spatial locations included in $M_{j}$, and $\sum_{m}$ denotes the summation over all the spatial locations in the attention map.
This energy function is optimized to maximize the correlation $A_{m,d}$ within the mask while minimizing the correlation outside of it.
Specifically, at each application of ControlNet for image restoration,  the gradient of the energy function (\ref{eq:ef}) is computed via backpropagation to update the latent $z_{j,t}$

\begin{equation}
 z_{j,t} = z_{j,t}- \eta \cdot \nabla_{z_{j,t}}E(A,M_j,k)
 \label{eq:bg}
 \end{equation}
 
where $\eta >0$ is a scale factor controlling guidance strength.

Meanwhile, to account for the preceding objects and their constraints during the restoration of the current object, we further combine the latent values of $z_{j, (t-1)}$ and $z_{(j-1), (t-1)}$. We fuse multiple object restorations during the diffusion sampling process in the latent space instead of pixel domain, so that no boundaries between objects would be introduced in the reconstructed image. Specifically, following the step $t$ in the reverse process (where $t$ transitions from its initial value to 0), we update the latent variable $z_{j,(t-1)}$ as follows:

\begin{equation}
    z_{j, (t-1)} = M_j \odot z_{j, (t-1)} + (1 - M_j) \odot z_{(j-1), (t-1)}
\end{equation}

where $\odot $ computes element-wise product. 
After $J$ iterations, we have successfully restored the detailed information for $J$ objects in the reference image. 

Finally, we utilize ControlNet to further restore the entire image given the overall description $Text_{all}$. This step plays a crucial role in the decoding process as it ensures consistency and enhances the overall perceptual quality of the entire image.

\begin{figure*}[t!]
    \centering
    \includegraphics[width=1.0\linewidth]{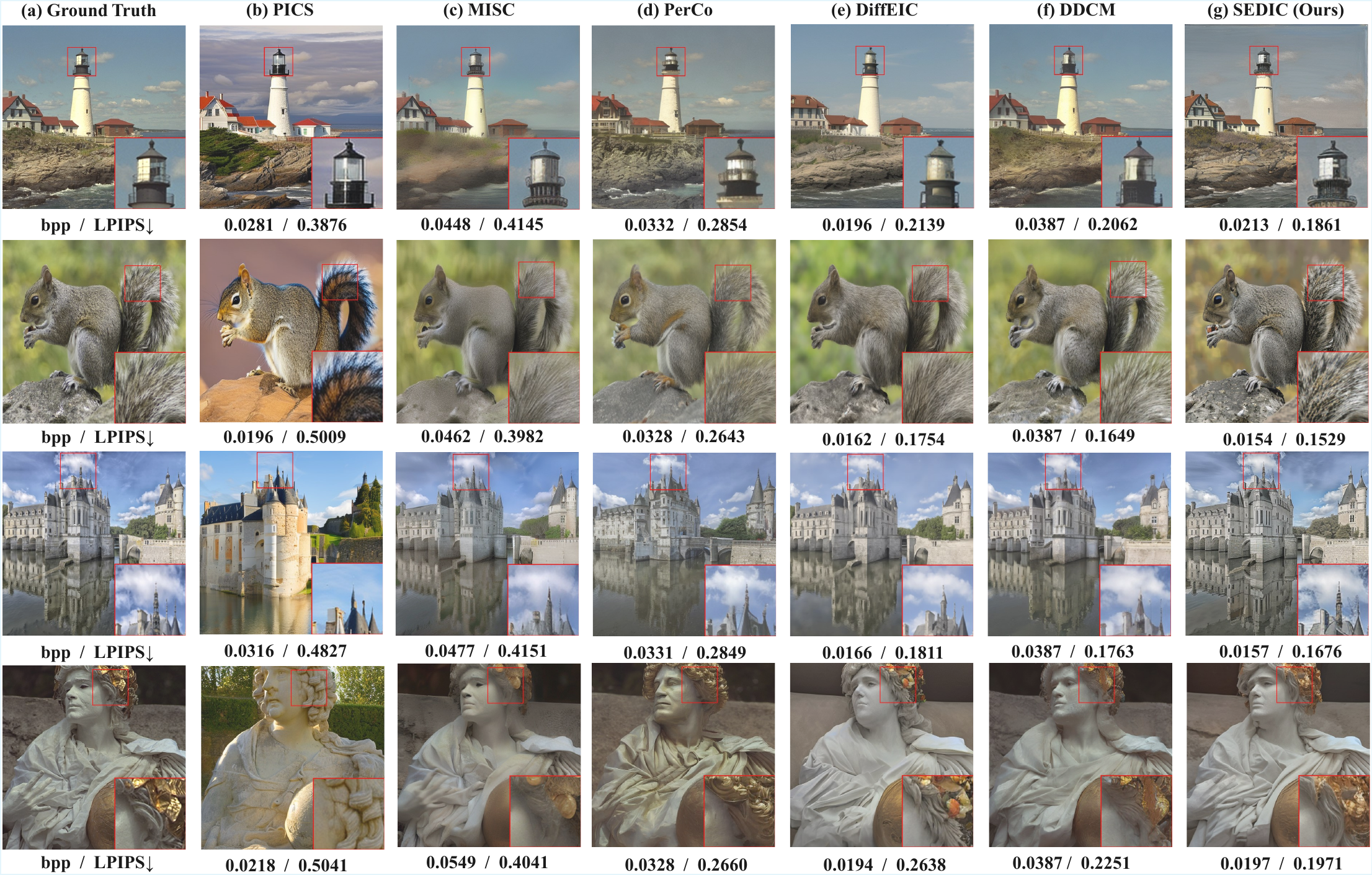}
\vspace{-3mm}    
    \caption{ We visually compare our SEDIC framework with stable diffusion-based methods on Kodak and DIV2K validation datasets under extremely low-bitrate settings. 
   The corresponding bpp and LPIPS values are displayed below the images.
}
\vspace{-3mm}
    \label{fig:vision_image}
\end{figure*}

\section{Experiment}

\subsection{Experimental Settings}

\noindent \textbf{Implementation:}
We keep the Image Textualization(GPT-4 Vision \cite{gpt4-vision}) and Semantic Mask Encoder(,Grounding Dino \cite{2023-groundingDino} and SAM \cite{2023-SAM}), along with ControlNet \cite{2023-controlnet}, frozen. Only an extremely low-bitrate image encoder/decoder is fine-tuned instead based on the cheng2020-attn model from the deep image compression platform CompressAI \cite{2020compressai}. Training begins at the lowest bitrate, with the loss weight scaled by reducing \( \lambda \) tenfold and a learning rate of $ 10^{-4} $.
Our SEDIC dynamically adjusts bitrates by tuning the number of objects $J$, word length of text descriptions $l_d$ and $l_{all}$. When $J$ is set to 1, with $l_d$ and $l_{all}$ designated as 20 and 30 words respectively, the bitrate falls within the range of 0.02 to 0.03 bpp. When $J$ increases to 3, with $l_d$ and $l_{all}$ designated as 30 and 50 words respectively, the bitrate is $0.04 \sim 0.05$ bpp. This relatively high bitrate allows for more image details and thus better recovery.
In the ORAG implementation, we adopt the middle block of the upsampling branch, as it provides the best trade-off between controllability and reconstruction fidelity \cite{chen2024training}. 
We found that hyperparameter $\eta$ between 30-50 work well across most settings and set $\eta = 40$ by default.

{
\setlength{\tabcolsep}{4pt}
\begin{table}[h]
    \centering
    \small
    \caption{Comparison of PSNR across different methods on Kodak, DIV2K validation, and CLIC2020 datasets. Each entry represents (PSNR $ \uparrow $, bpp).}
    \label{tab:psnr_bpp_comparison}
    \begin{tabular}{lccc}
        \toprule
        Method & Kodak  & DIV2K validation  & CLIC2020  \\
        \midrule
        PICS & (11.27, 0.0281) & (12.24, 0.0232) & (9.61, 0.0236) \\
        PerCo & (18.12, 0.0586) & (18.79, 0.0562) & (17.71, 0.0589) \\
        DiffEIC & (18.96, 0.0441) & (18.75, 0.0593) & (19.65, 0.0331) \\
        MISC& (20.21, 0.0438) & (19.81,0.0434) & (20.69,0.0452) \\
        DDCM& (21.56, 0.0564) & (20.37,0.0564) & (\textbf{23.14},0.0564) \\
        SEDIC & (\textbf{21.83}, 0.0457) & (\textbf{20.65}, 0.0546) & (22.72, 0.0439) \\
        \bottomrule
    \end{tabular} 
\end{table}
}

\begin{figure*}[t]
    \centering
    \includegraphics[width=1\linewidth]{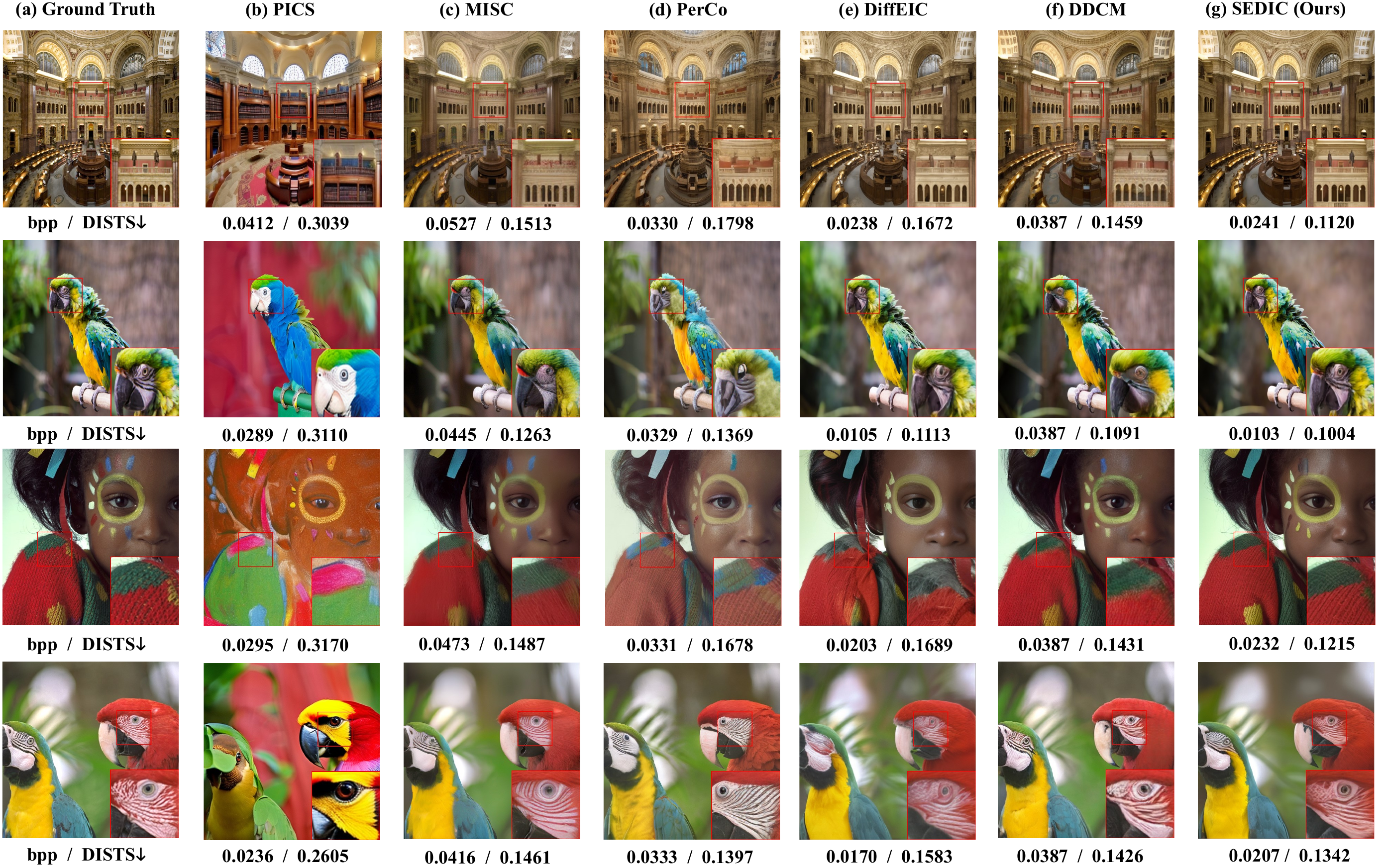}
    \vspace{-3mm}
    \caption{Visual comparison of the proposed SEDIC framework with Stable Diffusion-based methods on the Kodak and DIV2K datasets. For each method, the bpp and DISTS values are displayed below the images.}
    \label{fig:result_vision_DISTS}
    \vspace{-3mm}
\end{figure*}

\begin{figure*}[t]
    \centering
    \includegraphics[width=1\linewidth]{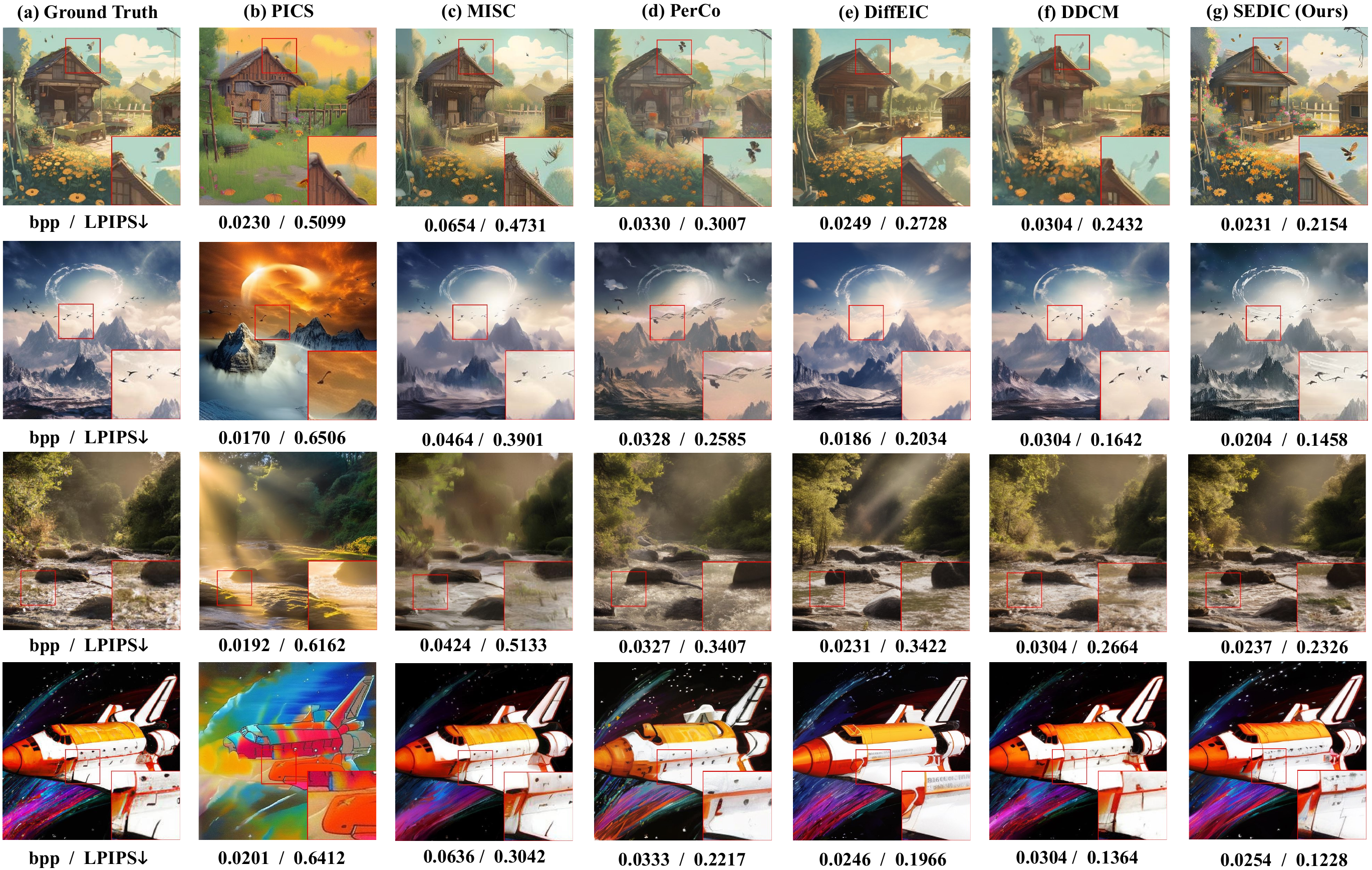}
    \vspace{-3mm}
    \caption{Visual comparison of the proposed SEDIC framework with Stable Diffusion-based methods on the AGIQA dataset. For each method, the bpp and LPIPS values are displayed below the images.}
    \label{fig:result_vision_AGIQA}
    \vspace{-3mm}
\end{figure*}

\noindent \textbf{Test Data:}
We evaluate on three standard benchmarks: Kodak~\cite{kodak} (24 natural images at 768×512), DIV2K validation~\cite{DIV2K} (100 images), and CLIC2020~\cite{clic2020} (428 images). For DIV2K and CLIC2020, images are resized to a minimum dimension of 768px and center-cropped to 768×768 for evaluation.

\noindent \textbf{Metrics:}
We adopt a comprehensive set of compression evaluation metrics to address both consistency and perceptual quality requirements. Perceptual metrics become crucial at extremely low-bitrates. They are prioritized over pixel-level metrics such as PSNR and SSIM. Specifically, we employ Learned Perceptual Image Patch Similarity (LPIPS) \cite{2018-LPIPS} and Deep Image Structure and Texture Similarity (DISTS) \cite{2020-DISTS} metrics to assess perceptual quality. We use standard no-reference metrics, Frechet Inception Distance (FID) \cite{2017-FID} and Kernel Inception Distance (KID) \cite{2018-KID}, to measure realism according to distributional alignment. In addition, ClipSIM \cite{2021-ICML-ClipSIM}, NIQE \cite{2012-SPL-NIQE}, and ClipIQA \cite{2023-AAAI-clipIQA} are also included to evaluate the semantic consistency between images. Additionally, the compression bitrate is assessed in terms of bits per pixel (bpp).

\begin{table}[t!]
    \centering
    \small
    \caption{More metrics results on the CLIC2020 dataset.}
    \label{tab:comparison}
    \begin{tabular}{lcccc}
        \toprule
        Method  & ClipSIM $\uparrow$ & NIQE $\downarrow$ & ClipIQA $\uparrow$ & bpp  \\
        \midrule
        PICS  & 0.8968 & 10.4208 & 0.6833 & 0.0236 \\
        PerCo  & 0.9291 & 10.9253 & 0.6741 & 0.0589 \\
        DiffEIC  & 0.9316 & 6.4063 & 0.6768 & 0.0331 \\
        MISC  & 0.9106 & 3.8271 & 0.6612 & 0.0470 \\
        DDCM  & 0.9367 & 3.3769 & 0.6843 & 0.0564 \\
        SEDIC (ours)  & \textbf{0.9630} & \textbf{3.2544} & \textbf{0.6917} & 0.0439 \\
        \bottomrule
    \end{tabular} 
    
\end{table}

\subsection{ Experiment Results and Discussion}

We compare our SEDIC with SOTA image compression methods, including traditional compression standards VVC \cite{2021-VCC}, BPG \cite{BPG}; learned image compression approaches MBT \cite{2018-MBT}, GAN based HiFiC \cite{2020-Hific}, Diffusion based approaches including CDC \cite{2024CDC}, PerCo \cite{2024-ICLR-PerCo}, DiffEIC \cite{2024-TCSVT-DiffEIC}, Mask image modeling based MCM \cite{2023-MCM} and Text-to-Image model based PICS \cite{lei2023text+sketch}, MISC \cite{li2024-misc}, DDCM \cite{2025-ICML-DDCM}. 
For VVC, we utilize the reference software VTM23.03 configured with intra-frame settings.

\begin{table}[t]
    \centering
    \small
    \caption{Ablation study on the effect of Attention Guidance(AG) in object restoration on CLIC2020 dataset. }
    \label{tab:ablation_soref}
    \begin{tabular}{lccccc}
        \toprule
        Method & LPIPS $\downarrow$ & DISTS $\downarrow$ & FID $\downarrow$ & KID $\downarrow$ & bpp \\
        \midrule
        w/ AG & \textbf{0.1756} & \textbf{0.1016} & \textbf{11.86} & \textbf{0.00162} & 0.0439 \\
        w/o AG & 0.2268 & 0.1427 & 15.77 & 0.00225 & 0.0439 \\
        \bottomrule
    \end{tabular}
    
\end{table}

\noindent \textbf{Quantitative Comparisons:}
Figure \ref{fig:result_image} presents the rate-distortion-perception curves of various methods on three datasets under extremely low-bitrate settings. It can be observed that our proposed SEDIC consistently outperforms SOTA compression approaches across all distortion and perception metrics, showing better semantic consistency and perceptual performance.
BPG \cite{BPG}, VVC \cite{vvc} and MBT\cite{2018-MBT} optimize the rate-distortion function in terms of MSE, leading to poor perception quality in terms of FID, DISTS and LPIPS. By contrast, Generative image compression approaches exhibits much better perception quality even at low bitrates, including HiFiC \cite{2020-Hific}, MISC \cite{li2024-misc}, PerCo \cite{2024-ICLR-PerCo}, DiffEIC \cite{2024-TCSVT-DiffEIC}, PICS \cite{lei2023text+sketch} and DDCM \cite{2025-ICML-DDCM}. 
Among these generative approaches, PICS \cite{lei2023text+sketch} encodes images into simple text and rough sketches, results in poor semantic consistency (higher LPIPS and DISTS) despite of high perception quality(low FID). DiffEIC \cite{2024-TCSVT-DiffEIC} becomes  SOTA baseline in terms of perception quality and semantic consistency. Our proposed SEDIC still outperforms SOTA baseline with a great margin. 

We compare the PSNR performance of different Diffusion-based methods on Kodak, DIV2K validation, and CLIC2020 datasets in Table~\ref{tab:psnr_bpp_comparison}. While SEDIC is primarily tailored for perceptual quality enhancement, it achieves the highest PSNR scores on the Kodak and DIV2K validation datasets, and performs competitively on the CLIC2020 benchmark, with PSNR marginally lower than that of DDCM. These results demonstrate that our framework effectively bridges the gap between perceptual optimization and traditional fidelity metrics through its novel coding mechanism, achieving a superior balance between perceptual quality and pixel-level fidelity. This enables SEDIC to deliver both visually appealing reconstructions and exceptional objective quality preservation.

To evaluate the semantic consistency and human perception performance of the methods, we conducted a comparative analysis of ClipSIM, ClipIQA, and NIQE on the CLIC2020 dataset. As shown in Table~\ref{tab:comparison}, our proposed SEDIC outperforms the others across all metrics. Specifically, SEDIC achieves the highest scores in ClipSIM (0.9630) and ClipIQA (0.6917), demonstrating its effectiveness in preserving semantic consistency and perceptual fidelity. Additionally, SEDIC attains the lowest NIQE score of 3.2544, further confirming its ability to improve the natural image quality. 

\begin{figure*}[t!]
    \centering
    \includegraphics[width=1.0\linewidth]{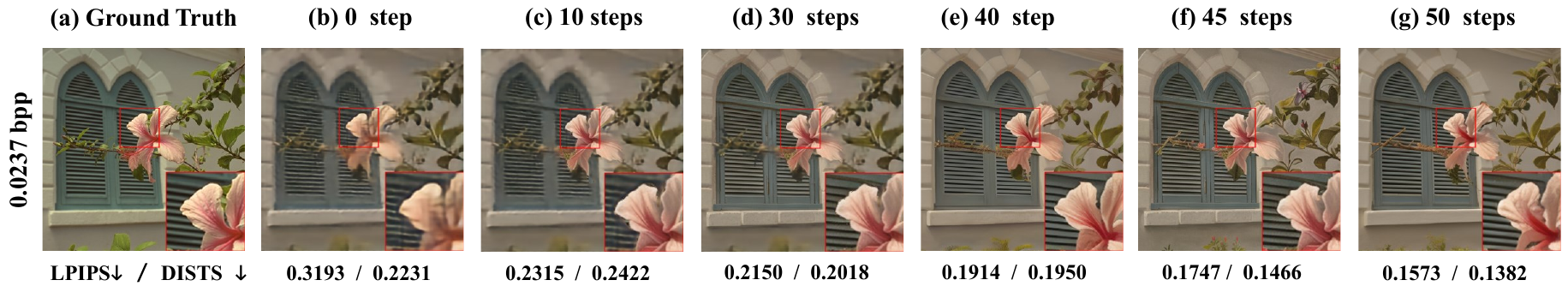}
    \vspace{-3mm}
    \caption{ Visual comparisons of different denoising steps. 0 step denotes the reference image as the starting point.}
    \vspace{-3mm}
    \label{fig:step_vision}
\end{figure*}

\begin{table*}[t!]
\centering
\small
\caption{Quantitative comparison on two datasets: Animals-10 (simple scene) and PASCAL VOC (complex scene). Lower is better for all metrics.}
\label{tab:animals10_pascalvoc_results}
\begin{tabular}{lccccccccccc}
\toprule
\multirow{2}{*}{Method} & \multicolumn{5}{c}{Animals-10 (Simple Scene)} & \multicolumn{5}{c}{PASCAL VOC (Complex Scene)} \\
\cmidrule(lr){2-6} \cmidrule(lr){7-11}
& LPIPS ↓ & DISTS ↓ & FID ↓ & KID ↓ & Bpp ↓ & LPIPS ↓ & DISTS ↓ & FID ↓ & KID ↓ & Bpp ↓ \\
\midrule
PICS    & 0.5756 & 0.3215 & 76.91 & 0.00924 & 0.0325 & 0.6823 & 0.5433 & 108.21 & 0.015927 & 0.0386 \\
PerCo   & 0.2729 & 0.1569 & 28.67 & 0.00532 & 0.0418 & 0.3419 & 0.4171 & 42.73  & 0.007914 & 0.0384 \\
DiffEIC & 0.0863 & 0.1643 & 23.71 & 0.00342 & 0.0458 & 0.3361 & 0.1952 & 37.81  & 0.004872 & 0.0414 \\
DDCM & 0.0931 & 0.1577 & 21.17 & 0.00245 & 0.0564 & 0.2738 & 0.1621 & 33.24  & 0.003611 & 0.0564 \\
SEDIC   & \textbf{0.0807} & \textbf{0.1491} & \textbf{20.64} & \textbf{0.00217} & 0.0426 
        & \textbf{0.2576} & \textbf{0.1571} & \textbf{32.78} & \textbf{0.003491} & 0.0405 \\
\bottomrule
\end{tabular}

\end{table*}

\noindent \textbf{Qualitative Comparisons:}
We visualize the visual quality performance of stable diffusion-based methods in Figure \ref{fig:vision_image} compared with PICS, MISC, PerCo, DiffEIC, and DDCM at extremely low-bitrates.  Notably, MISC exhibits limited ability to recover fine details of primary objects due to its weak guidance of object prompts to diffusion models. For example, the fur texture of the squirrel is poorly reconstructed. Compared to other methods, SEDIC achieves reconstructions with higher perceptual quality, fewer artifacts, and more realistic details at extremely low bitrates. For example, SEDIC preserves the fine details of the tower's peak that are lost or distorted in other methods (see the first row). Similarly, SEDIC generates more realistic fur details (e.g., the squirrel's tail in the second row).
Additionally, SEDIC better retains background cloud details (see the third row). In addition, Figure~\ref{fig:result_vision_DISTS} presents more visual examples at extremely low bitrates, along with the corresponding bpp/DISTS values. Furthermore, we evaluate the compression performance on the AI-generated image dataset AGIQA\footnote{https://github.com/lcysyzxdxc/AGIQA-3k-Database}
, using both subjective quality and the LPIPS metric. As shown in Figure~\ref{fig:result_vision_AGIQA}, our method achieves significant improvements over three competing approaches on this dataset. These results demonstrate that, compared with existing schemes, the proposed SEDIC algorithm is particularly effective for compressing AI-generated images. Overall, SEDIC attains higher perceptual quality, fewer artifacts, and more realistic detail reconstruction at such low bitrates. Notably, our method shows obvious advantages in restoring fine-grained object details, which can be attributed to the specially designed object restoration stage in our framework.

\begin{figure*}[t]
    \centering
    \includegraphics[width=1\linewidth]{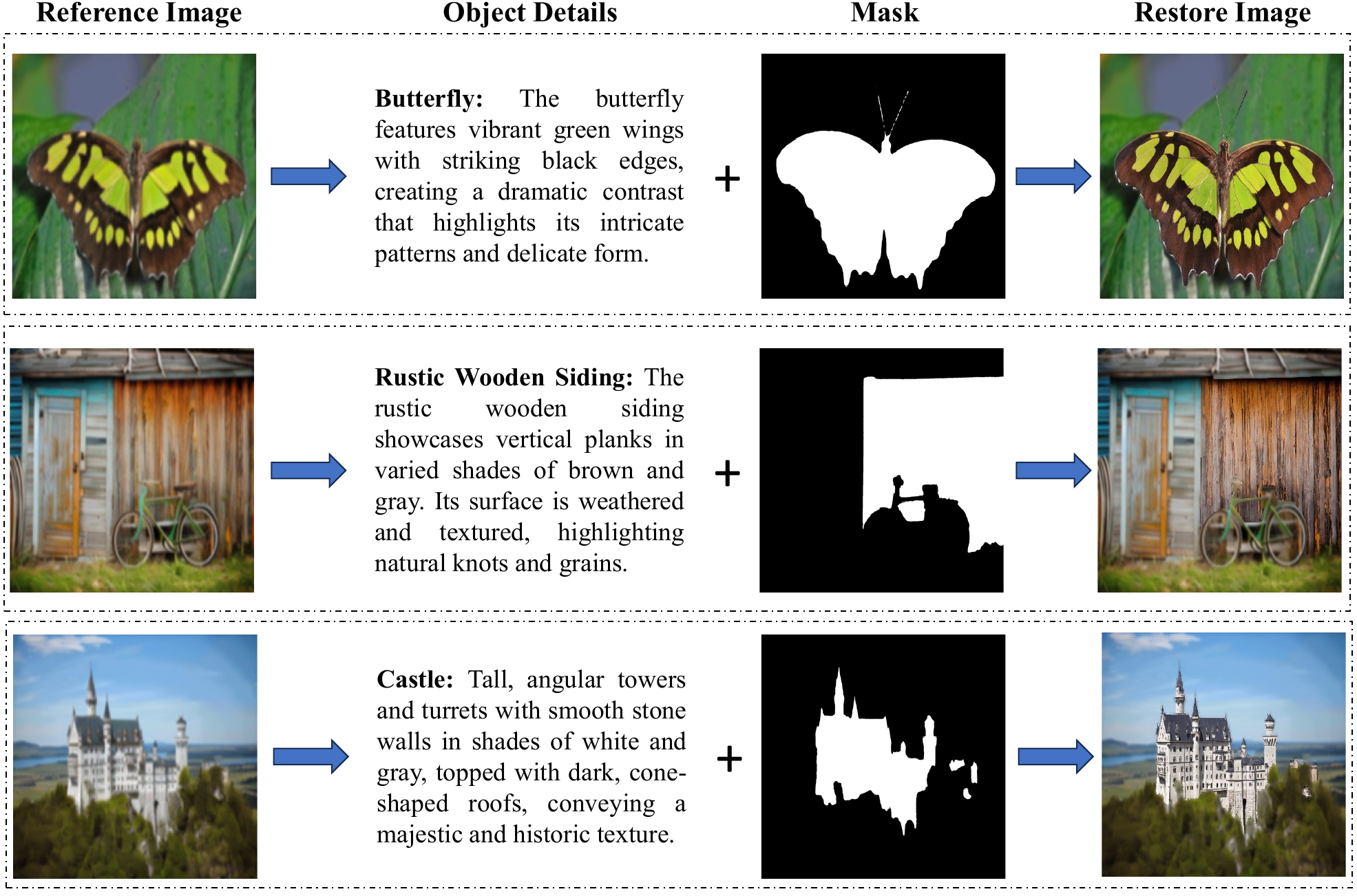}
    \caption{Visualization of the proposed SEDIC results in recovering object details .As shown, the object details are effectively restored by the proposed SEDIC.}
    \label{fig:text_object_step}
\end{figure*}

\begin{table}[t]
\footnotesize
\centering
\caption{Ablation results on Kodak~\cite{kodak}. $J$: number of objects; $Text_{all}$: overall image description; $\tilde{I}0$: reference image; $l_d$, $l{all}$: word lengths of object and overall descriptions.}
\label{table:ablation}
\setlength{\tabcolsep}{6pt}
\begin{tabular}{c|ccccc|c}
\toprule
\multicolumn{1}{c|}{\multirow{2}{*}{No}} & \multicolumn{5}{c|}{Content} & \multicolumn{1}{c}{\multirow{2}{*}{(LPIPS $\downarrow$, DISTS $\downarrow$, bpp)}} \\ \cmidrule{2-6}
& $J$ & $Text_{all}$ & $\tilde{I}_0$ & $l_d$ & $l_{all}$ & \\ 
\midrule
1 & 0 & \checkmark & \checkmark &     & 50  & (0.2338, 0.1667, 0.0226) \\
2 & 1 & \checkmark & \checkmark & 30  & 50  & (0.2260, 0.1522, 0.0304) \\
3 & 1 &            & \checkmark & 30  &     & (0.2517, 0.1760, 0.0258) \\
4 & 2 &            & \checkmark & 30  &     & (0.2327, 0.1641, 0.0334) \\
5 & 3 &            & \checkmark & 30  &     & (0.2243, 0.1503, 0.0412) \\
6 & 3 & \checkmark & \checkmark & 30  & 50  & (0.1518, 0.1012, 0.0457) \\
7 & 3 & \checkmark &            & 30  & 50  & (0.3518, 0.2284, 0.0275) \\
8 & 3 & \checkmark & \checkmark & 10  & 50  & (0.1718, 0.1318, 0.0413) \\
9 & 3 & \checkmark & \checkmark & 30  & 30  & (0.1651, 0.1151, 0.0442) \\
\bottomrule
\end{tabular}

\end{table}

\subsection{Experimental evaluation of simple and complex scenes}
Considering that object restoration may be related to the complexity of scenes, we evaluate our method on both simple (1-2 objects) and complex (5+ objects) scenes, using 100 test images from the Animals-10 and PASCAL VOC datasets.

As shown in Table~\ref{tab:animals10_pascalvoc_results}, our proposed method SEDIC consistently outperforms all existing approaches (PICS, PerCo, and DiffEIC) across both datasets. On the simple Animals-10 dataset, SEDIC achieves the lowest scores in all perceptual and realism metrics (LPIPS, DISTS, FID, and KID), demonstrating its superior reconstruction quality and visual fidelity in less cluttered scenes.
In the case of the complex PASCAL VOC dataset, although performance slightly degrades across all methods due to increased scene complexity, SEDIC maintains its leading position, again attaining the lowest scores in all key metrics. Notably, it significantly outperforms both PICS and PerCo, especially in terms of FID and KID, which measure realism and distributional similarity.

Overall, SEDIC demonstrates strong robustness and generalization ability across scenes of varying complexity, achieving a desirable trade-off between perceptual quality and bit-rate efficiency. Its advantages are particularly pronounced in simple scenes, while still maintaining competitive performance in more challenging, object-dense scenes.

\subsection{Object Restoration Visualization}
Figure~\ref{fig:text_object_step} illustrates the visual results of the proposed SEDIC framework in recovering fine-grained object details on the ultra-compressed reference image. Each object is restored under the guidance of its corresponding textual description and semantic mask. As shown in the figures, the object details are effectively restored, such as butterfly, Siding, castle, etc., demonstrating satisfactory perception quality and consistency. These examples highlight the effectiveness of our attention-guided object restoration strategy and its generalization across diverse semantic categories.

\subsection{Complexity Analysis}

We compare SEDIC with other compression methods in terms of computational complexity. Table~\ref{table:Complexity} reports the average encoding and decoding time (in seconds) on the Kodak dataset. 
Specifically, reference image encoding, mask generation, and text generation in our SEDIC framework take 0.054s, 0.117s, and 2.79s, respectively, which are all included in encoding time in Table~\ref{table:Complexity}.
It can be observed from the table that Diffusion-based methods generally incur higher computational cost than VAE- or GAN-based models.  Our SEDIC' encoding time is relatively longer than SOTA diffusion-based DiffEIC baseline due to text generation through GPT-4 Vision model. Notably, Our SEDIC still encodes much faster than PICS~\cite{lei2023text+sketch}, which requires iterative projection in the CLIP space for text generation. Our SEDIC's decoding time is comparable to PerCo and DiffEIC under equal denoising steps. As the denoising steps in the diffusion models increase, the decoding time increases dramatically. DDCM requires extensive denoising steps, which significantly increases its decoding time.

\subsection{Ablation Study}

We conducted an ablation study to evaluate the contribution of different semantically encoding components within SEDIC, as shown in Table \ref{table:ablation}. These components are designated as: 1) number of objects $J$, 2) Overall Description of the image $Text_{all}$, 3) extremely compressed reference image $\tilde{I}_0$, and 4) object description word length $l_d$ and overall description word length $l_{all}$. 
The results indicate that the extremely compressed reference image is the most essential component. Absence of the extremely compressed reference image brings dramatic perception quality degradation (Line 6 vs 7). Perceptual quality improves with more restored objects, highlighting the effectiveness of object-level semantic compression (Line~3~$\rightarrow$~5).
Additionally, the Overall Description also brings overall perception quality improvement during the decoding process (Line 3 vs 2). The word lengths of object descriptions $ l_d $ and overall descriptions $l_{all}$ have a slight impact on the results (Line6 vs Lines 8,9 ).

{
\setlength{\tabcolsep}{3pt}
\begin{table}[t]
\centering
\small
\caption{Encoding and decoding time (in seconds) on Kodak.}
\label{table:Complexity}
\begin{tabular}{cc|ccc}
\toprule
            Method & Step & Enc. Time(s) & Dec. Time(s) & Platform   \\\cmidrule{1-5} 
    VVC     & -   & 13.862 ± 9.821       &      0.066 ± 0.006                &   i9-13900K        \\
 HiFiC    &   -             &     0.038 ± 0.004                    &      0.059 ± 0.004                   &     RTX4090       \\ \cmidrule{1-5}    PICS    &   25         &    62.045 ± 0.516                     &    12.028 ± 0.413                     &   RTX4090        \\
                PerCo     &     20           &      0.080 ± 0.018                   &         2.551 ± 0.018                  &    A100        \\ 
                DiffEIC     &     20           &      0.128 ± 0.005                   &         1.964 ± 0.009                  &    RTX4090
        \\
                 DiffEIC     &     50           &      0.128 ± 0.005                   &          4.574 ± 0.006                   &    RTX4090
       \\
       DDCM     &     1000           &      0.172 ± 0.008                   &          24.613 ± 0.017                   &    RTX4090
       \\
     SEDIC(Ours)     &     20           &      2.947 ± 0.013                    &          2.332 ± 0.003                   &    RTX4090
       \\
     SEDIC(Ours)     &     50           &      2.947 ± 0.013                   &          4.994 ± 0.003                   &    RTX4090
       \\
\bottomrule 
\end{tabular}

\end{table}
}

Furthermore, to quantitatively assess the effect of attention guidance (AG) in object restoration on the reconstruction quality, we evaluate the performance in terms of LPIPS, DISTS, FID and KID metrics with and without attention guidance on CLIC2020 dataset. 
In the case of object restoration without attention guidance, we instead remove the attention guidance from Equation (\ref{eq:bg}). 
As shown in Table~\ref{tab:ablation_soref}, the attention backward guidance in object restoration has a significant impact on both the semantic consistency and perceptual quality of the reconstructed images. By incorporating this attention guidance, our decoder ensures that generated object details given by object descriptions are accurately positioned within the mask region. This backward attention mechanism contributes to more precise and visually coherent object restorations.

\begin{figure}[t!]
    \centering
    
    \includegraphics[width=1\linewidth]{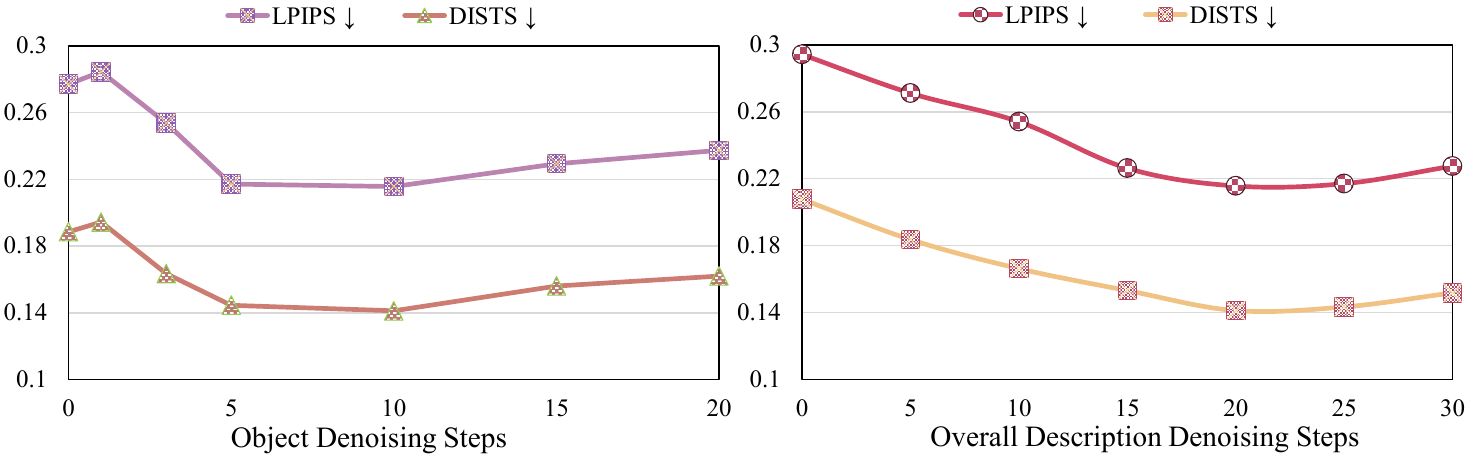}
    
    \caption{Quantitative comparison of reconstruction performance with different denoising steps for object-level and overall reconstruction on the Kodak dataset.
    }
    \label{fig:step_more_result}
\end{figure}

\begin{figure}[t!]
    \centering
    
    \includegraphics[width=0.95\linewidth]{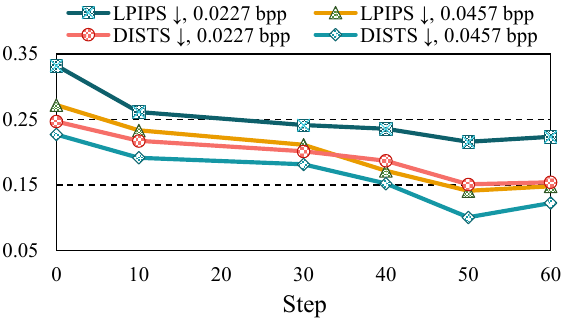}
   \vspace{-3mm}
    \caption{ Quantitative comparisons of different denoising steps on Kodak \cite{kodak}. 0 step denotes using reference image.}
    \label{fig:result_step}
  \vspace{-3mm} 
\end{figure}

\subsection{Effect of Denoising Steps}

Fig.\ref{fig:result_step} presents the reconstruction performance under varying denoising steps. We observe that increasing denoising steps generally enhances the perceptual quality of the decoded images. However, when denoising steps exceed 50, a slight degradation in quality is observed, suggesting that over-denoising may lead to detail loss. The diffusion-based decoder operates by first reconstructing object-level details from the extremely compressed reference image, followed by overall image refinement. All experiments are conducted with the object-level denoising steps fixed at 10. The visual results in Fig.\ref{fig:step_vision} further illustrate that more realistic and refined details emerge as the number of steps increases. 

Additionally, we conduct ablation studies to comprehensively investigate the allocation of denoising steps between object restoration and overall image restoration. First, we fixed the denoising steps to be 20 for overall restoration and evaluated the impact of varying denoising steps for object restoration. As shown in Figure \ref{fig:step_more_result} (left), the image perceptual quality improves as the Object denoising steps increase, reaching its peak around 10 steps. However, further increasing the denoising steps results in a slight degradation in reconstruction quality, possibly due to over-denoising, which may lead to fine detail loss.
Next, we fixed the object denoising steps to 10 and explored the influence of varying overall restoration denoising steps on reconstruction performance. As illustrated in Figure \ref{fig:step_more_result} (right), the best reconstruction results are achieved when the Overall Description Denoising Steps are set to around 20. This indicates that an appropriate number of steps effectively captures the global descriptive information while avoiding excessive smoothing or information loss. In summary, the experimental results demonstrate that both Object Denoising Steps and Overall Description Denoising Steps have a significant impact on image reconstruction quality. It is critical to properly balance the denoising steps between these two components.

\begin{figure}[t!]
    \centering
    
    \includegraphics[width=0.95\linewidth]{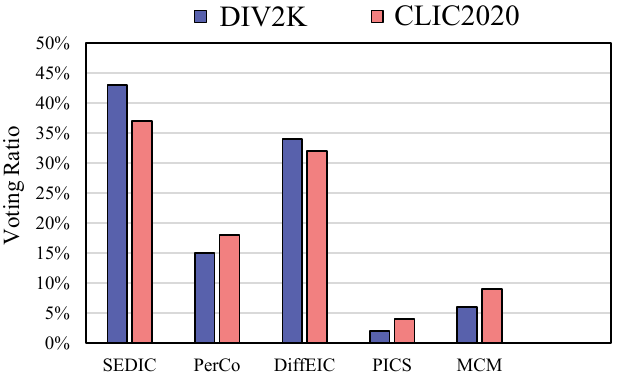}
    
    \caption{ Statistical results of user study in the CLIC-2020 and DIV2K datasets. Humans subjectively believe that our proposed SEDIC is the best compression metric for both consistency and perception.
}

    \label{fig:user_study}
\end{figure}

{
\setlength{\tabcolsep}{5pt}
\begin{table}[t]
    \centering
    \small
    \caption{Performance comparison on CLIC2020 dataset at comparable bitrates when using only overall descriptions and reference images to guide the diffusion model (J=0) and incorporating object-level details (J=3). }
    \label{tab:object_level_comparison}
    \begin{tabular}{lccccc}
        \toprule
        Method & LPIPS $\downarrow$ & DISTS $\downarrow$ & FID $\downarrow$ & KID $\downarrow$ & bpp \\
        \midrule
        SEDIC (J=0)   & 0.2939 & 0.1642 & 15.16 & 0.00263 & 0.0423 \\
        SEDIC (J=3)   & \textbf{0.1756} & \textbf{0.1016} & \textbf{11.86} & \textbf{0.00162} & 0.0439 \\
        \bottomrule
    \end{tabular}
    
\end{table}
}

{
\setlength{\tabcolsep}{4pt}
\begin{table}[t!]
    \centering
    \small
    \caption{Bitrate allocation of different encoding semantic components on Kodak, DIV2K validation, and CLIC2020 datasets.}
    \label{table:bpp_allocation}
    \begin{tabular}{lcccc}
        \toprule
        Dataset & Reference Image & Text  & Mask (J=3) & Total Bpp \\
        \midrule
        Kodak & 0.0181 & 0.0126 & 0.0152 & 0.0457 \\
        DIV2K val & 0.0228 & 0.0131 & 0.0187 & 0.0546 \\
        CLIC2020 & 0.0169 & 0.0127 & 0.0143 & 0.0439 \\
        \bottomrule
    \end{tabular}
      
\end{table}
}

\subsection{Bitrate Allocation Analysis }
Table \ref{table:bpp_allocation} presents bitrate (bpp) allocation for different semantic components across Kodak, DIV2K validation, and CLIC2020 datasets. The semantic components in our method consist of an extremely compressed reference image, text descriptions, and masks. The table demonstrates that the reference image occupies slightly more bitrates than text descriptions and masks. And it can be expected that more object items will consume more bitrates. This detailed analysis is crucial for understanding the role of each semantic component, which offers insights into the trade-offs between different semantic inputs.

\subsection{Effect of Object-level Restoration}

We intend to investigate which factor is more contributive in our proposed SEDIC  at a given bitrate: the quality of the compressed reference image itself, or the application of object restoration to a lower-quality reference.
We compare two variants on the CLIC2020 dataset at similar bitrates: (1) \( J=3 \), performing full multi-stage decoding from object to global restoration; (2) \( J=0 \), restoring the entire image using ControlNet conditioned only on the overall description and compressed reference.
As shown in Table~\ref{tab:object_level_comparison}, our proposed object restoration brings great performance gains compared to entire image restoration from the  compressed reference image only(J=0). Our proposed object restoration with attention guidance enables more specific restoration of object details, thereby enhancing the reconstruction quality of objects of interest. 

\subsection{User Study}
To validate the practicality of the proposed SEDIC in real-world scenarios, we conducted a subjective user study beyond objective metrics to analyze human preferences for compressed images. We randomly selected 100 images from the DIV2K and CLIC2020 datasets. Ten volunteers are invited to vote for the best reconstructed images based on consistency and perception quality among reconstructed images from different compression methods, including PerCo, DiffEIC, PICS, and MCM. For a fair comparison, we constrained the bitrates to $0.04 \sim 0.05$ bpp. The voting ratio results, shown in Figure \ref{fig:user_study}, demonstrate that the proposed SEDIC performs excellently across all assessed criteria, providing clear evidence of its effectiveness.

\section{Conclusion}

We propose a novel image compression framework SEDIC for extremely low-bitrate compression, which leverage LMMs to achieve extremely low-bitrate compression while maintaining high semantic consistency and perceptual quality. Specifically, the SEDIC approach leverages LMMs to Disentangl the images into compact semantic representations, including an extremely compressed reference image, overall and object-level text descriptions and the semantic masks. We propose an object restoration model with attention guidance, built upon the pre-trained ControlNet, to restore objects conditioned by the object detailed description and semantic masks. Based on that, we design a multi-stage decoder which performs restoration object by object progressively starting from the extremely compressed reference image, ultimately
generating high-quality and high-fidelity reconstructions. Extensive experimental results demonstrate that SEDIC significantly outperforms SOTA image compression methods in terms of perceptual quality at extremely low-bitrates($\le$ 0.05 bpp). We believe that this LMMs driven approach has the potential to pave the way for a new paradigm in image compression.

%

\bibliographystyle{IEEEtran}
\bibliography{ref}

\end{document}